\documentclass[runningheads]{llncs}
\usepackage{graphicx}
\usepackage{amsmath,amssymb} % define this before the line numbering.
\usepackage{color}
\usepackage[width=122mm,left=12mm,paperwidth=146mm,height=193mm,top=12mm,paperheight=217mm]{geometry}
\usepackage{float}

\usepackage[mathscr]{euscript}
\usepackage{subcaption}
\usepackage{algorithm}
\usepackage{algpseudocode}

\newcommand{\Loss}{\mathcal{L}}
\DeclareMathOperator*{\Exp}{\mathbb{E}}
\newcommand{\half}{\frac{1}{2}}
\newcommand{\conv}{\ast}
\newcommand{\hadamard}{\odot}

\newcommand{\NumEpochs}{150}

\begin{document}
% \linenumbers
\pagestyle{headings}
\mainmatter

\title{The Neural Painter: \\Multi-Turn Image Generation}
\titlerunning{The Neural Painter: Multi-Turn Image Generation}

\authorrunning{Benmalek \emph{et al.}}

\author{Ryan Y. Benmalek$^{1,4}$, Claire Cardie$^1$, Serge Belongie$^{1,2}$\\ Xiadong He$^3$, Jianfeng Gao$^4$}
\institute{$^1$Cornell University, $^2$Cornell Tech, $^3$JD AI Research, $^4$Microsoft Research\\ {\tt\small $^1$\{ryanai3, cardie\}@cs.cornell.edu, $^1$sjb344@cornell.edu, $^3$xiadong.he@jd.com, $^4$jfgao@microsoft.com }}

\maketitle

\begin{abstract}
In this work we combine two research threads from Vision/ Graphics and Natural Language Processing to formulate an image generation task conditioned on attributes in a multi-turn setting. By multi-turn, we mean the image is generated in a series of steps of user-specified conditioning information.
Our proposed approach is practically useful and offers insights into neural interpretability. We introduce a framework that includes a novel training algorithm as well as model improvements built for the multi-turn setting.
We demonstrate that this framework generates a sequence of images that match the given conditioning information and that this task is useful for more detailed benchmarking and analysis of conditional image generation methods. 
\keywords{Generative Adversarial Networks, Natural Language Processing, Vision, Generative Models, Interpretability, Recurrent Neural Networks}
\end{abstract}

\section{Introduction}
Generating high-quality images with arbitrary content based on user input, whether through natural language or a discrete user interface, has been a long-term goal and focal point of the Graphics and Vision community. In Natural Language Processing, a comparatively important goal has been that of building natural language interfaces for complex programs that interact with users to complete tasks collaboratively. For visual data, prior work has generated images from sentences \cite{cocogan,StackGAN,StackGANv2,AttnGAN} and attributes \cite{AGA,GenAttrController,CondGAN}. More complex areas such as video or unsupervised generation also exist \cite{VideoGan,DCGAN,GAN}. For language, examples include \cite{LearnParseDB,IntegratingPENLP,Back2Blocks,LatentPNCG,GenCode}; we point the reader to \cite{DialogForCode} for a more in-depth overview and motivation.
We define a task at the intersection of these two areas: \emph{ build a system that a user can interact with iteratively to generate images in a dynamic fashion, i.e., updating after every round of interaction}. Possible use cases for such a system range from a skilled designer rapidly prototyping and refining designs to an art hobbyist creating high quality illustrations using speech commands. At every round, the user provides input, and the system returns an image conditioned on the history of inputs so far.

\begin{figure}[tbh!]
  \centering
  \makebox[\textwidth]{\includegraphics[width=.9\paperwidth]{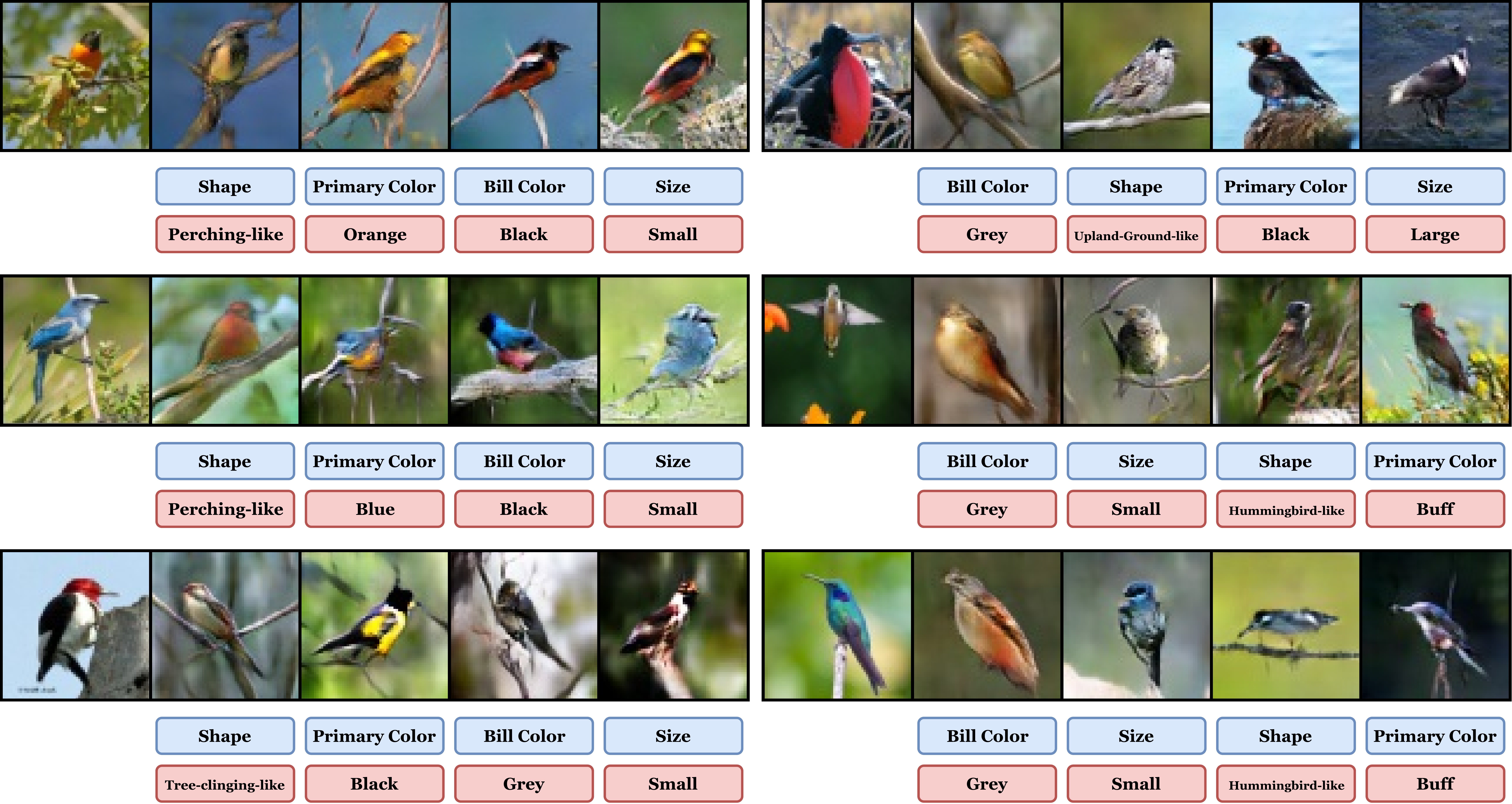}}
  \caption{Randomly selected validation set image sequences generated by our model. For each sequence, the ground truth image associated with the conditioning sequence is shown at far left. The sequence of images from left to right is generated using the associated context shown below. 
  }
  \label{fig:Seqs}
\end{figure}

Our proposed system also serves as a helpful testbed for debugging and improving conditional Generative Adversarial Networks (GANs) and benchmarking methods for interpretability in deep models. Our multi-turn setting disentangles the effects of different conditioning information on the rendered images, making statements like ``shape information is more difficult to integrate than color" quantifiable and explicit. We believe that such a setting can drive the community to find model improvements, modules, and training strategies to address specific problems in conditional GANs. We additionally justify this task as a setting for exploring neural interpretability by introducing the concept of visual justifications and arguing that they may be superior to current approaches to interpretability.

This framework presents new challenges that current methods cannot address. Because of the lack of supervision for the intermediate stages of image generation, this task is similar to episodic reinforcement learning, where the agent doesn't receive the reward until the end of the episode. In our case, we need to learn to produce reasonable intermediate images without direct supervision. This makes the current GAN-based approaches ill-suited since they implicitly assume a supervised case. There does exist a class of latent variable models like \cite{DRAW} that at first glance may seem to be sufficient since they update a `canvas' over the course of generation, but we note that these are latent variables and not human-comprehensible intermediate results, nor are they trained to be. Beyond the problem of intermediate supervision, applying current state of the art models in a non-recurrent approach has problems in not keeping images consistent through time. This applies both to attributes that are unrelated to the conditioning information (illumination, background, etc.) and for attributes that are related (pose, color).

In order to solve these challenges, we propose a novel model framework with a new training algorithm that allows us to hallucinate intermediate supervision without making strong distributional assumptions about conditioning or datasets used, and without requiring the existence of a sampler. This allows us to train our model to generate intermediate results without ever observing them. We display qualitative results showing that our model does learn to generate these intermediate results  responsively to conditioning information and coherently with the full sequence of generation, even when large changes are required. In doing so, we lay out a class of deep generative models that have recurrence through time, using primitive elements -- convolutional Gated Recurrent Units (GRUs) \cite{ConvGRU} -- originally designed for discriminative video based tasks. Additionally, we demonstrate concretely that this setting allows us to  understand better where and why the current state of the art in GANs fail at generating images, pointing to specific problems to tackle with the goal of improving image generation in general.

Our contributions in this work are as follows. We combine two research threads from two different areas -- vision/graphics and NLP -- in order to introduce a task that is practically useful and that offers insights into promising research directions in these areas and in conditional image generation and neural interpretability. We introduce a framework that includes a novel training algorithm as well as model improvements in order to handle the challenges introduced by his task. We demonstrate that this framework works well both in terms of the quality of images generated and in terms of performance on the task by presenting images as well as examples of specific changes occurring (shape, color, etc.).

\section{Related Work}
Beyond the brief overview of concrete problems and long-term goals in NLP that help to define our task, we would like to draw the reader's attention to the significant body of work in the Human-Computer Interaction community over the past several decades on the space of control, feedback, and multi-step interaction with computers and technology. Much of this work phrases these interactions as a form of collaborative conversation \cite{CollaborativeFeedback}, while also building upon empirical and theoretical results showing that people interact with computers in such a conversational fashion \cite{HCII} \cite{DisplayBasedAction}. The emphasis of \cite{DirectManipulationInterfaces} on incremental and rapid feedback to changes taken by the user suggest that our task has implications for building better experiences using vision and NLP, beyond the usefulness of phrasing problems as multi-turn for performance.

Since the introduction of GANs \cite{GAN}, there has been a surge of interest in image generation, from both the unconditional \cite{DCGAN} and the conditional perspectives. Work has explored generating images from captions \cite{cocogan,StackGAN,StackGANv2}, attributes \cite{AGA,GenAttrController}, as well as how to parametrize the models and training framework \cite{ACGAN}, beyond the original \cite{CondGAN}.

With regard to generation, perhaps the works most closely related to ours are \cite{DRAW} and \cite{DRAWCaption}. Both of these works are important to our work -- the former for the introduction of a mental `canvas' that a generator iteratively updates in order to produce a final image, and the latter for the use of such a model on a difficult language to image task. Our work is also closely related to \cite{GRAN}, although they do not update the image with conditioning information at each step, and do not enforce that the image should be complete at every step. We distinguish our framework by the requirement that intermediate generation be human-comprehensible, via an actual image and not a latent vector. 

\subsection{Neural Interpretability}
To date, the majority of the work on interpretability in neural networks can be broadly categorized either as post-training methods, which find neurons selective for specific concepts or objects \cite{ObjectDetectors}\cite{NetworkDissection} or as attention-based methods, which provide internal attention maps to illustrate where the network is focusing as it makes predictions.

Although these methods do provide a more intuitive view into understanding the operation of these networks on single examples and on datasets as a whole, they are not applicable on all tasks, particularly in complex settings like VQA \cite{VQA}  where multiple modalities are involved, and fail to meaningfully communicate what models are `thinking' at a high level. In contrast, we propose to solve the neural interpretability problem as a problem of visual justification -- where the model must explain its actions at every step by providing visual output, and where people can, at a glance, understand what the model is `thinking'.

This is in the same spirit as newer work which provides natural language justifications for model predictions in sentiment analysis \cite{Rationalizing}, and image classification \cite{GeneratingVisualExplanations}. We go a step further, creating visual justifications that illustrate the model's internal representation, not just its final actions, and our visual justifications are amenable to cases where a model must justify its `mental process' during computation as opposed to waiting till the very end. Accordingly, we expect to generate images at every turn of the conversation demonstrating the bot's current estimate of the image from the conditioning information so far and that these images should meaningfully change through the course of the conversation as new information is obtained (e.g. `primary color: red')

\section{Framework}
\subsection{Problem Definition}
Our task, simply posed, is as follows: the image generation model participates in a ``conversation" with an external actor who provides conditioning information at every turn that the generator must use to produce images. For example, this information takes the form of attribute-value pairs of a bird in question. Note that our framework applies to more complex cases: e.g. where the conditioning information at each turn may be a full sentence. 

\subsection{Dataset} In contrast to \cite{cocogan,StackGAN,StackGANv2}, and many of the GANs operating on the CUB dataset, which use a class-disjoint train/test split of the data, we conduct a $90/10$ stratified sample of the CUB dataset by class. 
In order to clearly illustrate our setting and to avoid the problem of dealing with very fine-detailed attribute classes, we restrict the attributes used to the top 4 observed in the dataset by frequency.

The attributes selected are the 4 most common in the dataset: ``Primary Color," ``Shape," ``Size," and ``Bill Color." These were chosen to provide good coverage over the types of attributes we could use, while containing both very disjoint (e.g., ``Shape" and ``Bill Color"), as well as less disjoint (e.g., ``Primary Color", ``Bill Color") attribute class pairs allowing us to observe how these affect training and generation. In addition, this choice helps us to disentangle the problem of vocabulary sparsity from our setting, in a first step toward what we want in our models: dealing with highly diverse and sparse vocabularies that characterize fine-grained changes across long timescales in a highly interacting fashion.

\section{Model}
\subsection{Overview}
Our model consists of two components: a reader submodule, to properly integrate conditioning information through time, and a recurrent generator, to transform this integrated information into pictures at each time step. Because of the importance of pretraining embeddings for GANs in terms of image quality and convergence, we pretrain these components separately, and then tune them together.
\subsection{Reader}
\begin{figure}[]
  \centering
  \makebox[\textwidth]{\includegraphics[width=.75\paperwidth]{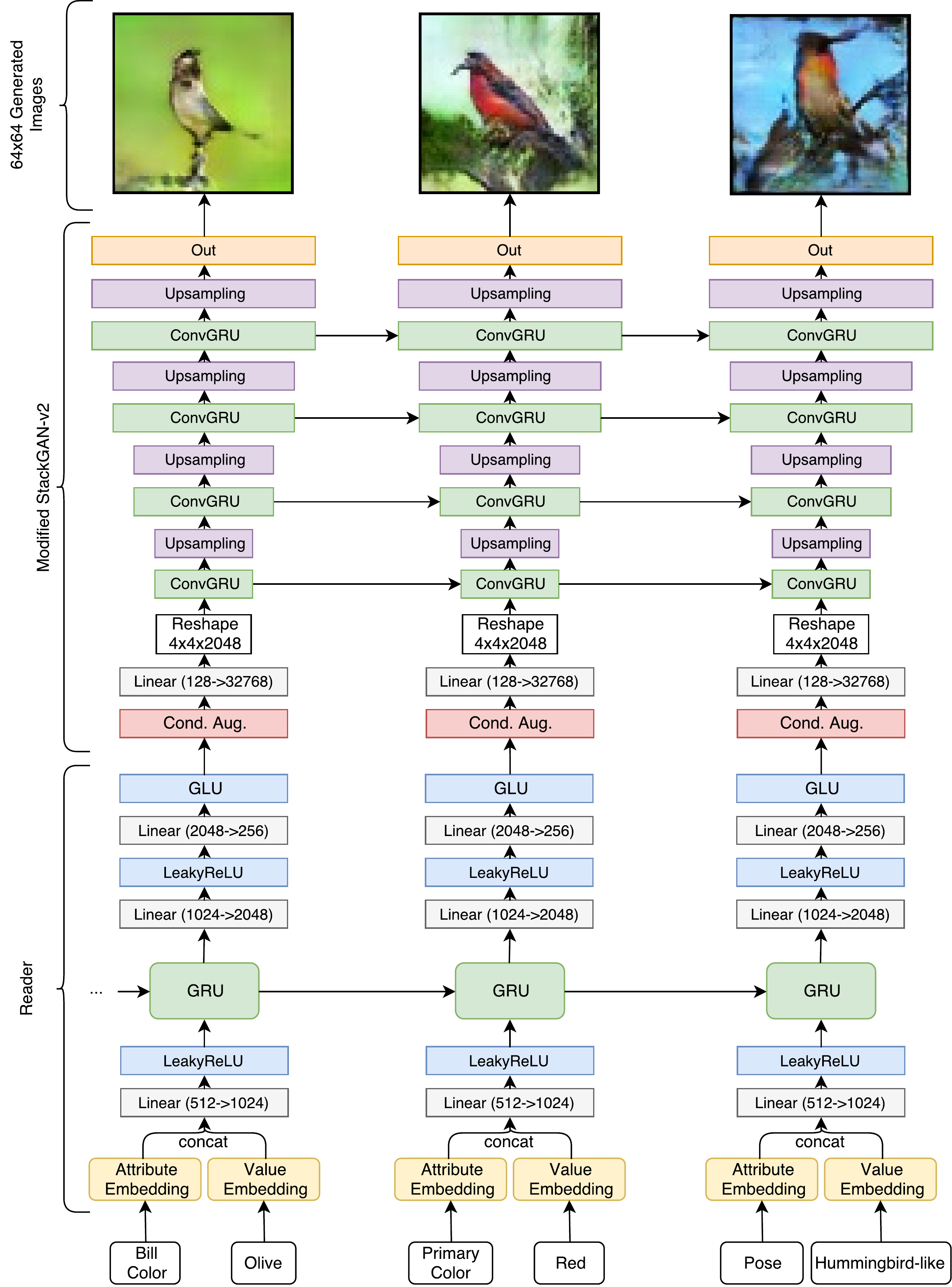}}
  \caption{
    Diagram of our model, demonstrating the output of the model on a validation set instance of length 3 and the conditioning information at bottom that generated it. Note that the images generated reflect current and past attribute-value pairs with high fidelity. After reading the pair ``Primary Color: Red"  the bird is now red, and after ``Pose: Hummingbird" the bird retains the primary red coloration while the shape changes and white coloration on the chest is added -- two more indicative traits of a hummingbird. The generated image sequence outlined in this figure was randomly chosen from the first run of the model on the validation set}
  \label{fig:Model}
\end{figure}

We first embed attributes and values, encode them together with a linear layer, and utilize a GRU to manage the relevant state updates through time as we receive new inputs. A nonlinear stack of transformations follows, as specified in Fig.~\ref{fig:Model}. Intuitively, we want our reader to observe conditioning information at each step and return embeddings that are useful for generating images that match the given input.

\subsection{StackGAN++ and modifications}
\subsubsection{Conv-GRU and recurrence in generation}
We use a convolutional GRU as introduced in \cite{ConvGRU}. With the convolutional modifications, the Conv-GRU equations are as follows:
\begin{equation}\label{eq:ConvGRU}
\begin{aligned}
  z_t & = \sigma(W_z \conv [h_{t-1},\ x_t]) \\
  r_t & = \sigma(W_r \conv [h_{t-1},\ x_t]) \\
  \tilde{h}_t & = \tanh(W \conv [r_t \hadamard h_{t-1}, \ x_t]) \\
  h_t & = (1 - z_t) \hadamard h_{t-1} + z_t \hadamard \tilde{h}_t
\end{aligned}
\end{equation}
where $\conv$ and $\hadamard$ denote convolution and the Hadamard product, respectively. \\

Intuitively, adding recurrence in the generator can help with training, by removing some of the burden of keeping state through time and integrating new information that falls on the reader module, as well as improving performance by giving the network the ability to condition on past generation - in the same way that \cite{StackGAN} \cite{StackGANv2} \cite{LapGAN} and \cite{ProgGAN} demonstrated that conditioning on lower-resolution generation can improve higher-resolution image generation. Finally, the use of recurrence increases the ability of the network to keep elements not related or not closely related to the conditioning information like pose, illumination, background, etc. fixed through the sequence of images. \\

We use a GRU recurrence instead of an LSTM recurrence in order to reduce training times, as well as parameter sizes and memory/computational requirements. The system was not sensitive to this difference. We demonstrate the first use of this class of modules in a non-discriminative context. Beyond this brief overview, we point the reader to \cite{ConvGRU} and \cite{ConvLSTM} to learn more about these convolutionally-recurrent models and how they can improve performance.

\subsubsection{Conditional Augmentation}
As shown in Fig. \ref{fig:Model}, the history of attribute-value pairs is first encoded by a reader module, yielding an embedding $c_t$ for each turn $t$. We use the conditional augmentation in \cite{StackGANv2} to improve training and diversity, where the output of the conditional augmentation layer is drawn from an independent Gaussian Distribution $\mathcal{N}(\mu(c_t), \Sigma(c_t))$ where the mean $\mu(c_t) \in \mathbb{R}^{100}$ and diagonal covariance matrix $\Sigma(c_t)$ are functions of the reader's output $c_t$. We additionally concatenate noise $z \sim \mathcal{N}(0, 1)$ which is fixed through time.

\subsection{Hyperparameters}
For the GAN component, we use the same hyperparameters as Stackgan++ \cite{StackGANv2}. All convolutional GRUs have kernel size 1 and have the same number of channels as the layer they run on top of. For the reader module, the GRU has hidden size 1024 and all other hyperparameters are described in the model diagram.

\section{Learning}
\begin{figure}[tbh!]
  \begin{subfigure}{\textwidth}
  \centering
    \includegraphics[width=.8\linewidth]{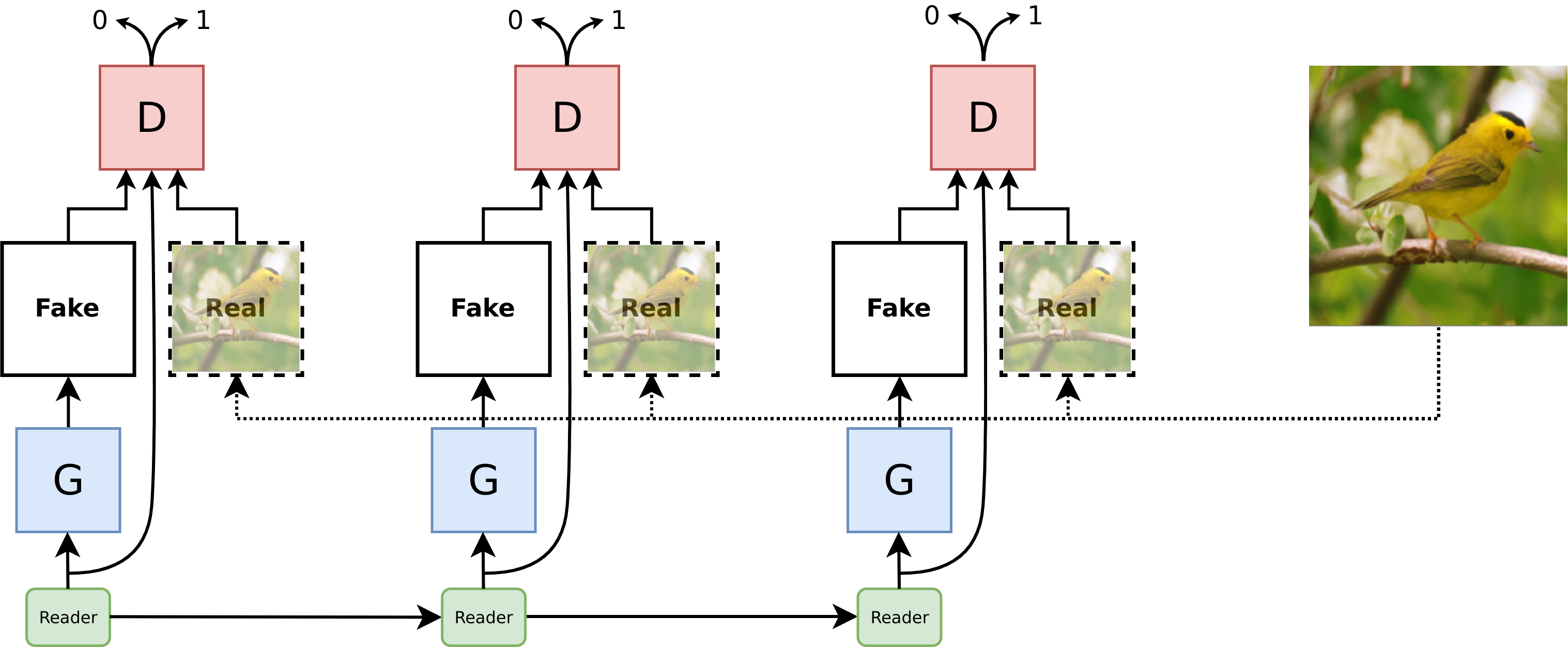}
    \caption{The naive multi-step approach to training}
    \label{fig:UnifSampSub1}
  \end{subfigure}
  \begin{subfigure}{\textwidth}
  \centering
    \includegraphics[width=.8\linewidth]{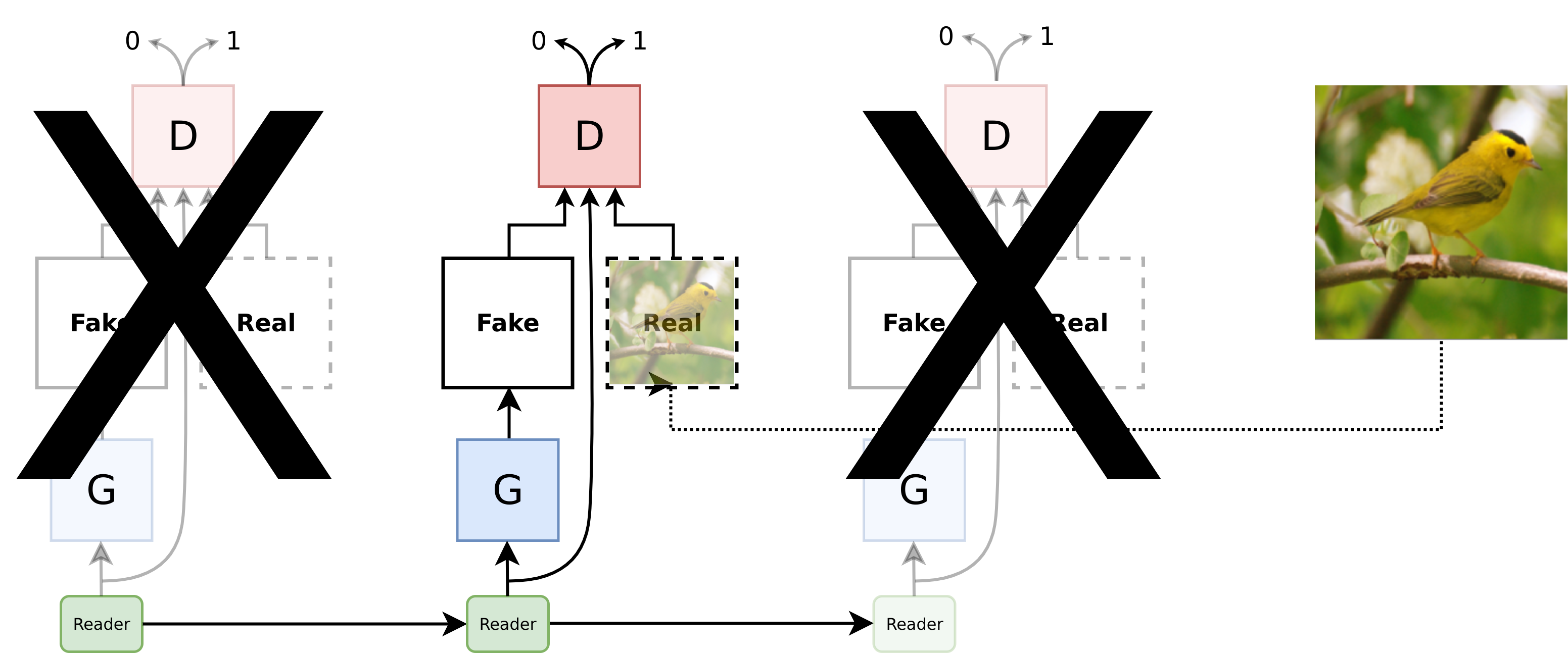}
    \caption{Our Uniform Sampling approach for $t = 1$}
    \label{fig:UnifSampSub2}
  \end{subfigure}
  \caption{Contrasting our approach to training with the naive approach. Instead of using the image $I$ as supervision for every round, we randomly sample a $t \sim Unif(0, T)$ and only backprop through the GAN for that step in the sequence. In this way we avoid the local optimum of generating highly similar, unchanging images from step to step.}
  \label{fig:UnifSamp}
\end{figure}

In order to modify the joint conditional-unconditional losses for the Discriminators and Generator, from \cite{StackGANv2} for the multi-turn setting, we need to find supervision for an image $I_t$ at turn $t$ in the conversation.
\begin{equation}\label{eq:LD-gold}
\begin{aligned}
  \Loss_D &= \frac{1}{T}\sum_{t=0}^{T}(
    \underbrace{
      -\half \Exp_{I_t \sim \mathcal{I}(c_t)}[\log(D(I_t))] \;\;\;\;\;
      -\half \Exp_{I_g \sim G(c_t)}[\log(1-D(I_g))]
    }_\text{unconditional loss} + \\
    &\;\;\;\;\;\;\;\;\;\;\;\;\;\;\;
    \underbrace{
      -\half \Exp_{I_t \sim \mathcal{I}(c_t)}[\log(D(c_t, I_t))] \;
      -\half \Exp_{I_g \sim G(c_t)}[\log(1-D(c_t, I_g))]
    }_\text{conditional loss}
  )
\end{aligned}
\end{equation}
\begin{equation}\label{eq:LG-gold}
\begin{aligned}
  \Loss_G &= \frac{1}{T}\sum_{t=0}^{T}(
    \Exp_{I_g \sim G(c_t)}[
      \underbrace{
        \half (1 - D(I_g))
      }_\text{unconditional loss} + 
      \underbrace{
        \half (1 - D(c_t, I_g))
      }_\text{conditional loss}
    ]
  )
\end{aligned}
\end{equation}
where $I_g$ is the image generated by the generator, $T$ is the length of conversations, $c_t$ is the history of the conversation from time $0$ to $t$, and $\mathcal{I}(c_t)$ is the conditional distribution of images that match $c_t$. Unfortunately, naively setting $I_t = I$ for all $t \in [0, \ n\_rounds)$, and training with this loss for all rounds in a batch tends to produce unchanging image sequences of lower quality. See Fig. \ref{fig:UnifSamp} for an illustration of our approach in contrast to the naive approach. \\
$G(c_t)$ defines a distribution over images sampled from by picking $z \sim \mathcal{N}(0, 1)$ resulting in $G(concat(c_t, z))$. 
For ease of notation, we define the following:
\begin{equation}\label{eq:LD_defn}
\begin{aligned}
  & L_D^R(I_t, c_t) = \half(\log D(x) + \log D(I_t, c_t)) \\
  & L_D^G(c_t) = \half \Exp_{I_g \sim G(c)}[\log(1-D(I_g)) + log(1-D(I_g, c_t))] \\
  & L_G(c_t) = \half \Exp_{I_g \sim G(c)}[(1-D(I_g)) + (1-D(I_g, c_t))] \\
  & L_D(I_t, c_t) = L_D^R(I_t, c_t) + L_D^G(c_t) \\
%  & L_D^T(x, c) = \frac{1}{T}\sum_{t=0}^T L_D(x, c_t) = \Exp_{t \sim Unif(T)}[L_D(x, c_t)]
\end{aligned}
\end{equation}
We would like to approximate the following loss:

\begin{equation}\label{eq:FubiniTrick}
\begin{aligned}
  \Exp_{c \sim \mathcal{C}}\left[\frac{1}{T}\sum_{t=0}^T(\Exp_{I \sim \mathcal{I}(c_t)} L_D(I, c_t))\right] = \Exp_{c \sim \mathcal{C}} \Exp_{t \sim Unif(T)} \Exp_{I \sim \mathcal{I}(c_t)} L_D(I, c_t)
\end{aligned}
\end{equation}
where $\mathcal{C}$ is the distribution over conversations (i.e. the distribution over conditioning strings of length $T$). For problems where the conditioning information is discrete and has a small vocabulary, it is possible to get a good approximation to the conditional distribution $\mathcal{I}(c_t)$ by sampling from the dataset, but this method will not work well for datasets that are sparse in this conditional distribution. To illustrate: the likelihood of the first two sentences describing two different images being exactly the same is vanishingly small for any reasonably-sized vocabulary, even when these images are semantically similar. To work on more complex problems where this data sparsity exists, we need a method that doesn't require sampling from this conditional distribution. \\ 
By Fubini-Tonelli's theorem, we can swap the first two expectations,
\begin{equation}\label{eq:PostSwap}
\begin{aligned}
  \Exp_{t \sim Unif(T)} \Exp_{c \sim \mathcal{C}} \Exp_{I \sim \mathcal{I}(c_t)} L_D(I, c_t)
\end{aligned}
\end{equation}
we then collapse the last two expectations into:
\begin{equation}\label{eq:FinalUnifLoss}
\begin{aligned}
  \Exp_{t \sim Unif(T)} \Exp_{ (c, I) \sim \mathcal{CI}} L_D(I, c_t)
\end{aligned}
\end{equation}
where $\mathcal{CI}$ is the joint distribution over images and conditioning information.
A similar argument follows for $L_G$. This loss function admits a simple learning algorithm; see Alg. \ref{alg:TrainAlg}.

\begin{figure}[t!]
\begin{tabular}{@{}p{7cm}}
\vspace{-0.275in}
\begin{algorithm}[H]
\caption{
  Hallucinating Supervision via Time-Uniform Sampling.
  We use the default values $\alpha_D = \alpha_G = 0.0002$, $T = 4$, $E = \NumEpochs$.
  }\label{alg:TrainAlg}
\begin{algorithmic}[1]
  \footnotesize
  \Require 
    $\alpha_G, \alpha_D$, the learning rate for the generator and discriminator, respectively. 
    $m$, the batch size. 
    $T$, the number of turns in the sequence. 
    $E$, the number of epochs to run. 
    $\theta_G$ and $\theta_D$, the initial generator and discriminator's parameters, respectively.
%  \State \Comment{This is how you comment}
  \For{$e=1$ to $E$, batch $\{ (c^{(i)}, I^{(i)})\}_{i=1}^m$ in epoch}
    \State Sample $t \sim Unif(T)$
    \State Run reader over each $c^{(i)}_{0..t}$ to obtain $r^{(i)}_{0..t}$
    \State Run generator over $r^{(i)}_{0..t}$ to obtain $I^{(i)}_{f, 0..t}$
    \State $g_D \gets \nabla_D[\frac{1}{m}\sum_{i=1}^m L_D(I^{(i)}_t, r_t)$], using $I^{(i)}_{f, t}$ for the $L_D^G$ term.
    \State $\theta_D \gets \theta_D + \alpha_D \cdot \text{Adam}(\theta_D, g_D)$
    \State $g_G \gets \nabla_G[\frac{1}{m}\sum_{i=1}^m L_G(c_t)]$ using $I^{(i)}_{f, t}$ for the $L_G$ term.
    \State $\theta_G \gets \theta_G - \alpha_G \cdot \text{Adam}(\theta_G, g_G)$
  \EndFor
\end{algorithmic}
\label{alg:learn} 
\end{algorithm}
\end{tabular}
\end{figure}

\subsection{Pretraining}
We use the associated image embeddings $y_{sg}$ for StackGAN++ \cite{StackGANv2}, and train our reader module to take in conditioning information and predict the embedding at every step under mean squared error for 15 epochs using Adam with default parameters and a batch size of 32. Reader training was not sensitive to these hyperparameters. 

We also attempted to use embeddings from a variety of networks pretrained on imagenet \cite{Imagenet} as well as networks pretrained for CUB classification, discovering that image quality is highly sensitive to the embeddings used for pretraining, and the class-conditional loss introduced by \cite{cocogan} used to train $y_{sg}$ was important to learning high quality image embeddings that translate well to training a GAN.

We pretrain the stackgan component (generator and discriminator) without the conv-gru layers to generate images using $y_{sg}$ for 150 epochs, using all the hyperparameters from \cite{StackGANv2} without the KL-Divergence regularization they introduce, and without dropping the learning rate as they do.

We then ``put the models together" by initializing the full model with the parameters from the pretrained reader, generator, and discriminator, and train for $\NumEpochs$ epochs, keeping the optimizer hyperparameters the same, except we drop the learning rate by a factor of $2$ every $50$ epochs. We train with a batch size of 64 and perform simple dataset augmentation via random horizontal flips. Note that the Conv-GRU layers in the model  are initialized from scratch using the standard pytorch \cite{pytorch} initialization. 

\section{Results and Analysis}

\subsection{Common Changes}
We illustrate 4 common changes in generated images we see as a result of updating from given conditioning information: Color, Shape, Size, and Part. Part refers to changes localized to a specific part of the bird (most typically the bill), while Shape and Size changes require more intelligent adjustment of the generated image to remain coherent with the history so far.
\begin{figure}[h!]
  \centering
  \makebox[\textwidth]{\includegraphics[width=.8\paperwidth]{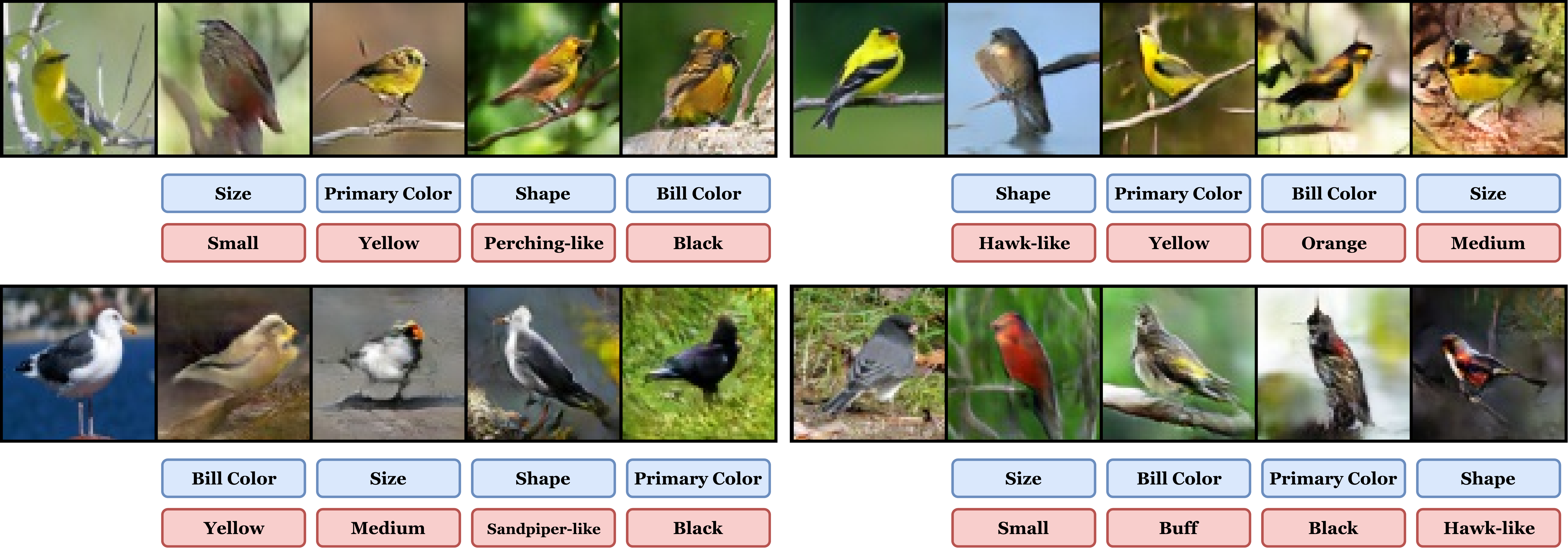}}
  \caption{We observe both simple and complex color changes in accordance with the conditioning information given. In the top two examples listed, the primary color and bill color change abruptly after receiving the conditioning information and remain for the rest of the sequence.}
  \label{fig:Color}
\end{figure}
\begin{figure}[tbh!]
  \centering
  \makebox[\textwidth]{\includegraphics[width=.85\paperwidth]{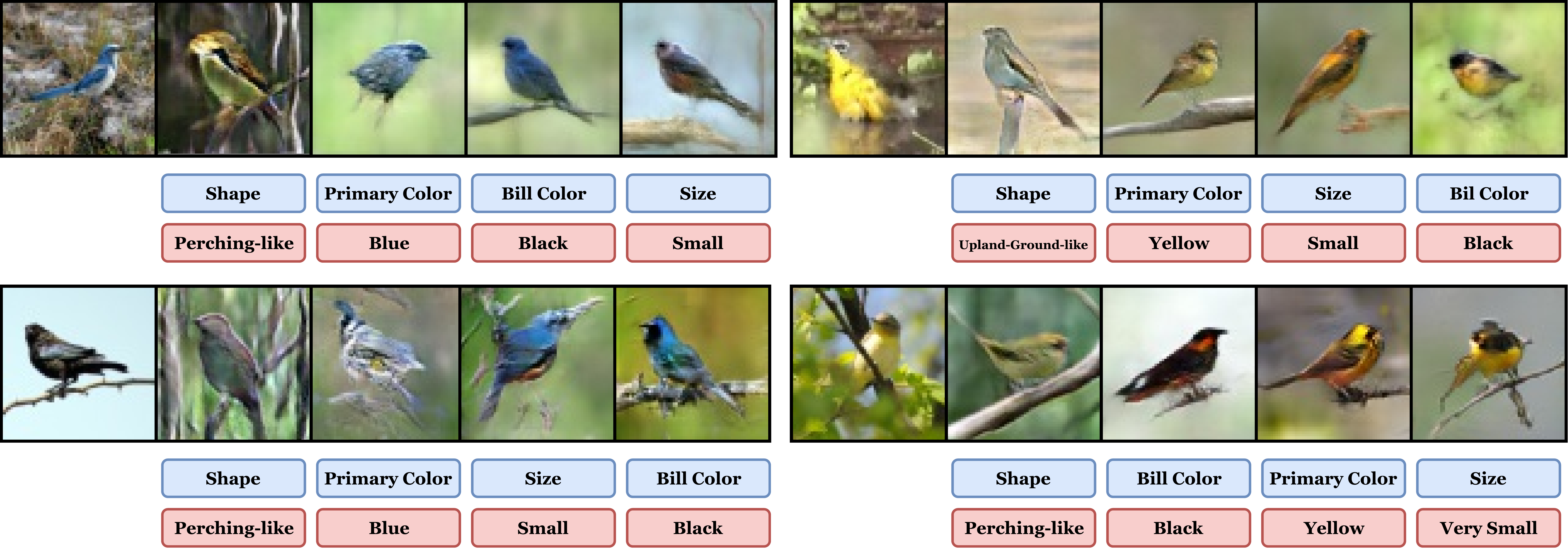}}
  \caption{We observe changes to the color of specific parts of the bird (primarily the bill, as in bottom left and top right) that persist through the conversation and at times modulate a more complex color pattern (bottom right)}
  \label{fig:Parts}
\end{figure}
\begin{figure}[h!]
  \centering
  \makebox[\textwidth]{\includegraphics[width=.85\paperwidth]{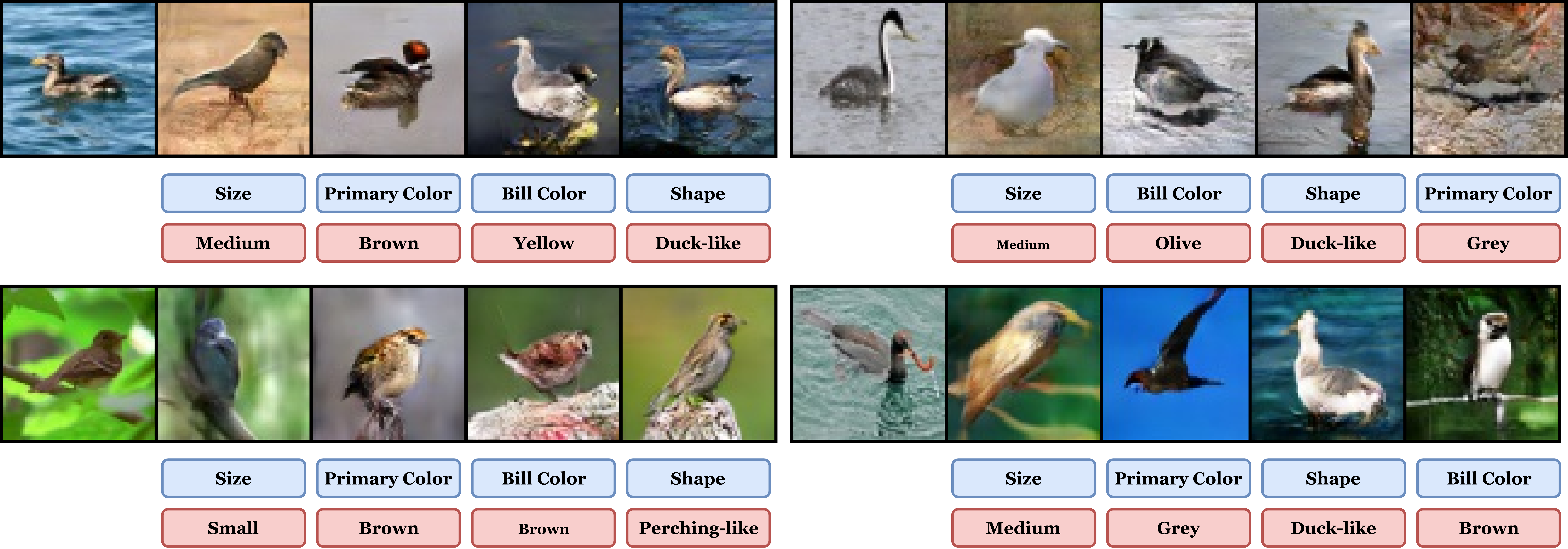}}
  \caption{Note that the model's response to conditioning information about shape not only affects the shape of the bird generated, but also the background of the scene (as in the bottom right where the background changes from sky to water). This suggests the network is learning more complex associations between conditioning information and components of an image. In the bottom left we see an example of a change in pose that is mostly invariant to other characteristics of the bird and the scene, illustrating the ability of the network to keep track of important factors through the full conversation.}
  \label{fig:Shape}
\end{figure}
\begin{figure}[h!]
  \centering
  \makebox[\textwidth]{\includegraphics[width=.85\paperwidth]{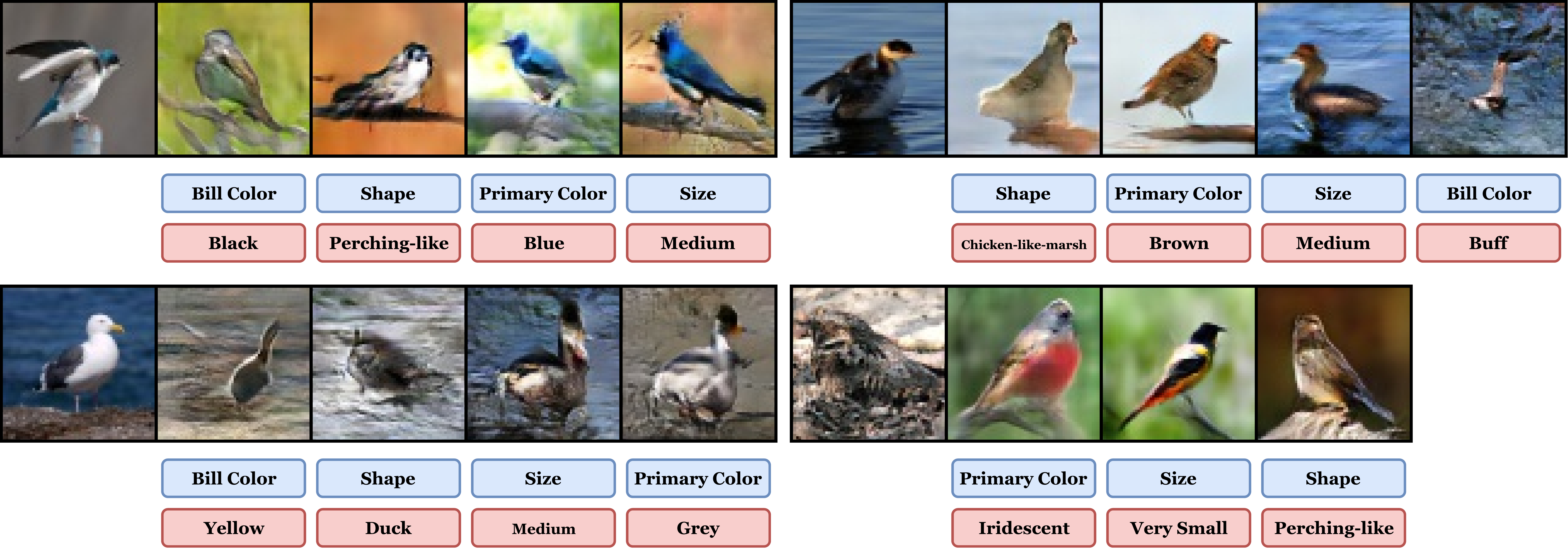}}
  \caption{We observe size conditioning information being used both to shrink or grow an already generated bird (top left, bottom right), and to clarify the species of the bird in question (bottom left, top right). This demonstrates that our model is learning size as both an operator that alters the currently generated image and a source of information for generating images during the rest of the conversation.}
  \label{fig:Size}
\end{figure}

\section{Conclusions}
In this paper we introduce a novel multi-turn image generation task where the model is asked to generate images at every step conditioned on a sequence of input information. A key challenge here is the lack of intermediate supervision during training, which we overcome with a novel learning algorithm and model improvements. Our learning algorithm hallucinates intermediate supervision in a provably equivalent way to training on the intermediate supervision if it existed. More broadly, this task is a helpful testbed for debugging and improving conditional GANs and benchmarking methods for interpretability in deep models.

\section*{Acknowledgements}
This work was initiated during an internship at Microsoft Research.
We thank Andrew Bennett, Valts Blukis, Daniel Jiwoong Im, Dipendra Misra, and Ayush Sekhari for helpful discussions.

%\par\vfill\par

%\clearpage

\bibliographystyle{splncs}
\bibliography{paper}
\end{document}